\documentclass[10pt,twocolumn,letterpaper]{article}

\usepackage{iccv}
\usepackage{times}
\usepackage{epsfig}
\usepackage{graphicx}
\usepackage{amsmath}
\usepackage{amssymb}
\usepackage{enumitem}
\usepackage[ruled,linesnumbered]{algorithm2e}

\usepackage[pagebackref=true,breaklinks=true,letterpaper=true,colorlinks,bookmarks=false]{hyperref}

 \iccvfinalcopy 

\def\iccvPaperID{170} 

\ificcvfinal\fi
\begin{document}


\title{DeepID-Net: multi-stage and deformable deep convolutional neural networks \\for object detection }

\author{ Wanli Ouyang, Ping Luo, Xingyu Zeng, Shi Qiu, Yonglong Tian, Hongsheng Li, Shuo Yang, \\
Zhe Wang, Yuanjun Xiong, Chen Qian, Zhenyao Zhu, Ruohui Wang,\\
 Chen-Change Loy, Xiaogang Wang, Xiaoou Tang 
\\
the Chinese University of Hong Kong\\
{\tt\small {wlouyang, xgwang}@ee.cuhk.edu.hk}
}

\maketitle

\begin{abstract}
In this paper, we propose multi-stage and deformable deep convolutional neural networks for object detection. This new deep learning object detection diagram has innovations in multiple aspects. In the proposed new deep architecture, a new deformation constrained pooling (def-pooling) layer models the deformation of object parts with geometric constraint and penalty. With the proposed multi-stage training strategy, multiple classifiers are jointly optimized to process samples at different difficulty levels. A new pre-training strategy is proposed to learn feature representations more suitable for the object detection task and with good generalization capability. By changing the net structures, training strategies, adding and removing some key components in the detection pipeline, a set of models with large diversity are obtained, which significantly improves the effectiveness of modeling averaging. The proposed approach ranked \#2 in ILSVRC 2014. It improves the mean averaged precision obtained by RCNN, which is the state-of-the-art of object detection, from $31\%$ to $45\%$. Detailed component-wise analysis is also provided through extensive experimental evaluation. 

\end{abstract}

\section{Introduction}
Object detection is a one of the fundamental challenges in computer vision.  It has attracted a great deal of research interest \cite{Dalal:HOG,Smeulders:SelectiveSearch,LatSVMObj}. The main challenges of this task are caused by the intra-class variation in appearance, lighting, backgrounds,  and deformation. 
In order to handle these challenges, a group of interdependent components in the pipeline of object detection are important. 
First, features should capture the most discriminative information of object classes. Well-known features include hand-crafted features such as Haar-like features \cite{Viola:ped}, SIFT \cite{Lowe:SIFT}, HOG \cite{Dalal:HOG}, and learned deep CNN features \cite{sermanet2013overfeat, Krizhevsky:ImageNetCNN, girshick2014rich}.
Second, deformation models should handle the deformation of object parts, e.g. torso, head, and legs of human. The state-of-the-art deformable part-based model (DPM) in \cite{LatSVMObj} allows object parts to deform with geometric constraint and penalty. 
Finally, a classifier decides whether a candidate window shall be detected as enclosing an object. SVM \cite{Dalal:HOG}, Latent SVM \cite{LatSVMObj}, multi-kernel classifiers \cite{Vedaldi09}, generative model \cite{mikolajczyk2006multiple}, random forests \cite{Dollar:Crosstalk}, and their variations are widely used. 

In this paper, we propose multi-stage deformable DEEP generIc object Detection convolutional neural NETwork (DeepID-Net). In DeepID-Net, we learn the following key components: 1) feature representations for a large number of object categories, 2) deformation models of object parts, 3) contextual information for objects in an image. We also investigate many aspects in effectively and efficiently training and aggregating the deep models, including bounding box rejection, training schemes, objective function of the deep model, and model averaging. The proposed new diagram significantly advances the state-of-the-art for deep learning based generic object detection, such as the well known RCNN \cite{girshick2014rich} framework. With this new pipeline, our method ranks \#2 in object detection on the ImageNet Large Scale Visual Recognition Challenge (ILSVRC) 2014. This paper also provides detailed component-wise experimental results on how our approach can improve the mean Averaged Precision (AP) obtained by RCNN \cite{girshick2014rich} from 31.0\% to mean AP 45\% step-by-step on the ImageNet object detection challenge validation 2 dataset.

The contributions of this paper are as follows:
\begin{enumerate}[leftmargin=12pt,noitemsep,nolistsep]
\item A new deep learning diagram for object detection. It effectively integrates feature representation learning, part deformation learning, sub-box feature extraction, context modeling, model averaging, and bounding box location refinement into the detection system.
\item A new scheme for pretraining the deep CNN model. We propose to pretrain the deep model on the ImageNet image classification dataset with 1000-class object-level annotations instead of with image-level annotations, which are commonly used in existing deep learning object detection \cite{girshick2014rich}. Then the deep model is fine-tuned on the ImageNet object detection dataset with 200 classes, which are the targeting object classes of the ImageNet object detection challenge. 
\item A new deformation constrained pooling (def-pooling) layer, which enriches the deep model by learning the deformation of visual patterns of parts. The def-pooling layer can be used for replacing the max-pooling layer and learning the deformation properties of parts at any information abstraction level.
\item We show the effectiveness of the multi-stage training scheme in generic object detection. With the proposed deep architecture, the classifier at each stage handles samples at a different difficult level. All the classifiers at multiple stages are jointly optimized. The proposed new stage-by-stage training procedure adds regularization constraints to parameters and better solves the overfitting problem compared with the standard BP.
\item A new model averaging strategy. Different from existing works of combining deep models learned with the same structure and training strategy, we obtain multiple models by using different network structures and training strategies, adding or removing different types of layers and some key components in the detection pipeline. Deep models learned in this way have large diversity on the 200 object classes in the detection challenge, which makes model averaging more effective. It is observed that different deep models varies a lot across different object categories. It motivates us to select and combine models differently for each individual class, which is also different from existing works \cite{zeiler2013visualizing, sermanet2013overfeat, he2014spatial} of using the same model combination for all the object classes. 
\end{enumerate}

\section{Related Work}
It has been proved that deep models are potentially more capable than shallow models in handling complex tasks \cite{Bengio:Deep}. Deep models have achieved spectacular progress in computer vision \cite{Hinton:DBN, Hinton:DBNs, ranzato2007unsupervised, jarrett2009best, Le:DBNLargeScale, Norouzi:CRBM2009,Krizhevsky:ImageNetCNN, zeiler2011adaptive,Luo:DeepFace, Sun2013Hybrid, Farabet2013Learning,Poon2011SPN}. Because of its power in learning feature representation, deep models have been widely used for object recognition and object detection in the recent years \cite{sermanet2013overfeat, zeiler2013visualizing,he2014spatial,Simonyan14a,zou2014generic, gong2014multi, lin2013network, girshick2014rich}. In existing deep CNN models, max pooling and average pooling are useful in handling deformation but cannot learn the deformation penalty and geometric model of object parts. The deformation layer was first proposed in our earlier work \cite{Ouyang2013JointDeep} for pedestrian detection. In this paper, we extend it to general object detection on ImageNet. In \cite{Ouyang2013JointDeep}, the deformation layer was constrained to be placed after the last convolutional layer, while in this work the def-pooling layer can be placed after all the convolutional layers to capture geometric deformation at all the information abstraction levels. All different from \cite{Ouyang2013JointDeep}, the def-pooling layer in this paper can be used for replacing all the pooling layers. In \cite{Ouyang2013JointDeep}, it was assumed that a pedestrian only has one instance of a body part, so each part filter only has one optimal response in a detection window. In this work, it is assumed that an object has multiple instances of a body part (e.g. a car has many wheels), so each part filter is allowed to have multiple response peaks in a detection window. This new model is more suitable for general object detection. 

Since some objects have non-rigid deformation, the ability to handle deformation improves detection performance. Deformable part-based models were used in \cite{LatSVMObj, Zhulong:StructSVM, Park:MultResObjDet, Ouyang2013MultiPed} for handling translational movement of parts. To handle more complex articulations, size change and rotation of parts were modeled in \cite{PictorialStruct}, and mixture of part appearance and articulation types were modeled in \cite{Bourdev:Poslet, Yang:Articulated, Desai:PhraseletsECCV12}. In these approaches, features are manually designed, Deformation and features are not jointly learned.

%
%
%
The widely used classification approaches include various boosting classifiers \cite{Dollar:Crosstalk,Dollar:ChnFtrs, Wu:edgelet}, linear SVM \cite{Dalal:HOG}, histogram intersection kernel SVM \cite{Maji:HikSvm}, latent SVM \cite{LatSVMObj}, multiple kernel SVM \cite{Vedaldi:MultKernel}, structural SVM \cite{Zhulong:StructSVM}, and probabilistic models \cite{Barinova:DetHough, Mikolajczyk:objdetGen}. In these approaches, classifiers are adapted to training data, but features are designed manually. If useful information has been lost at feature extraction, it cannot be recovered during classification. Ideally, classifiers should guide feature learning.

Researches on visual cognition, computer vision and cognitive neuroscience have shown that the ability of human and computer vision systems in recognizing objects is affected by the contextual information like non-target objects and contextual scenes. 
The context information investigated in previous works includes regions surrounding objects \cite{Dalal:HOG, Ding:ContextPedDet,galleguillos2010multi}, object-scene interaction \cite{Divvala:ContextObjDet}, and the presence, location, orientation and size relationship among objects \cite{Barinova:DetHough, Wu:IJCV07,Yan:PedDet,Desai:ObjLayout, Park:MultResObjDet, galleguillos2010multi, Song:ObjDetContext,Divvala:ContextObjDet,Yao:posObj,Ding:ContextPedDet,Yang:Proxemics,ouyang2013DBN2Ped, Desai:PhraseletsECCV12, sadeghi2011recognition, tang2013learning}. 
In this paper, we utilize the image classification result from the deep model as the contextual information.

In summary, previous works treat the components individually or sequentially. This paper takes a global view of these components and is an important step towards joint learning of them for object detection.

\section{Dataset overview}
The ImageNet Large Scale Visual Recognition Challenge (ILSVRC) 2014 \cite{ILSVRCarxiv14} contains two different datasets: 1) the classification and localization dataset and 2) the detection dataset. 

The classification and localization (Cls-Loc) dataset is split into three subsets, train, validation (val), and test data. The train data contains 1.2 million images with labels of $1,000$ categories. The val and test data consist of $150,000$ photographs, collected from flickr and other search engines, hand labeled with the presence or absence of $1,000$ object categories. The $1,000$ object categories contain both internal nodes and leaf nodes of ImageNet, but do not overlap with each other. A random subset of $50,000$ of the images with labels are used as val data and released with labels of the $1,000$ categories. The remaining $100,000$ images are used as the test data and are released without labels at test time.  The val and test data does not have overlap with the train data.

The detection (Det) dataset contains 200 object categories and is split into three subsets, train, validation (val), and test data, which separately contain $395,918$, $20,121$ and $40,152$ images. The manually annotated object bounding boxes on the train and val data are released, while those on the test data are not. The train data is drawn from the Cls-Loc data. In the Det val and test subsets, images from the CLS-LOC dataset where the target object is too large (greater than 50\% of the  image area) are excluded. Therefore, the Det val and test data have similar distribution. However, the distribution of Det train is different from the distributions of Det val and test. For a given object class, the train data has extra negative images that does not contain any object of this class. These extra negative images are not used in this paper. We follow the RCNN \cite{girshick2014rich} in splitting the val data into val$_1$ and val$_2$. Val$_1$ is used for training models while val$_2$ is used for validating the performance of models. The val$_1$/val$_2$ split is the same as that in \cite{girshick2014rich}.

\section{Method}
\subsection{The RCNN approach}
\label{Sec:RCNN}
\begin{figure} 
\centering
\centerline{\epsfig{figure=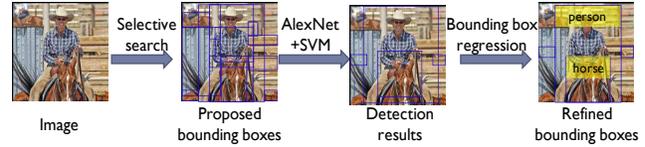,width=8.5cm}}
\caption{Overview of RCNN in \cite{girshick2014rich}. Selective search \cite{Smeulders:SelectiveSearch} is used for proposing candidate bounding boxes that may contain objects. AlexNet is used to extract features from the cropped bounding box regions. Based on the extracted features, SVM is used to decide the existence of objects. Bounding box regression is used to refine bounding box location and reduce localization errors. }
\label{Fig:ModelRcnn}
\vspace{-10pt}
\end{figure}

A brief description of the RCNN approach is provided for giving the context of our approach. RCNN uses the selective search in \cite{Smeulders:SelectiveSearch} for obtaining candidate bounding boxes from both training and testing images. An overview of this approach is shown in Fig. \ref{Fig:ModelRcnn}.

At the testing stage, the AlexNet in \cite{Krizhevsky:ImageNetCNN} is used for extracting features from bounding boxes, then 200 one-versus-all linear classifiers are used for deciding the existence of object in these bounding boxes. Each classifier provides the classification score on whether a bounding box contains a specific object class or not, e.g. person or non-person. The bounding box locations are refined using the AlexNet in order to reduce localization errors.
 
At the training stage, the ImageNet Cls-Loc dataset with $1,000$ object classes is used to pretrain the AlexNet, then the ImageNet Det dataset with $200$ object classes is used to fine-tune the AlexNet. The features extracted by the AlexNet are then used for learning 200 one-versus-all SVM classifiers for 200 classes. Based on the features extracted by the AlexNet, a linear regressor is learned to refine bounding box location.

\subsection{Overview of the proposed approach}

\begin{figure} 
\centering
\centerline{\epsfig{figure=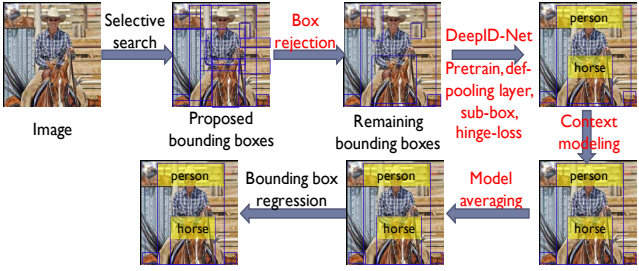,width=8.5cm}}
\caption{Overview of DeepID-Net. Selective search is used for proposing candidate bounding boxes that may contain objects. Then RCNN is used for rejecting $94\%$ candidate bounding boxes. Each remaining bounding box goes through the DeepID-Net in order to obtain 200 detection scores. Each score measures the confidence on whether the bounding box contains a specific object class, e.g. person, or not. After that, context is used for refining the 200 scores of each bounding box. Model averaging and bounding box regression are then used to improve the accuracy. Texts in red highlights the steps that are not present in RCNN \cite{girshick2014rich}.}
\label{Fig:ModelAll}
\vspace{-10pt}
\end{figure}

An overview of our proposed approach is shown in Fig. \ref{Fig:ModelAll}. In this model: 
\begin{enumerate}[leftmargin=12pt,noitemsep,nolistsep]
\item  The selective search in \cite{Smeulders:SelectiveSearch} is used for obtaining candidate bounding boxes. Details are given in Section \ref{Sec:RegionSel}.
\item An existing detector is used for rejecting bounding boxes that are most likely to be background. Details are given in Section \ref{Sec:RegionRej}. 
\item The remaining bounding boxes are cropped and warped into $227\times 227$ images. The $227\times 227$ cropped image goes through the DeepID-Net in order to obtain 200 detection scores. Each detection score measures the confidence on the cropped image containing one specific object class, e.g. person. Details are given in Section \ref{Sec:DeepIdNet}.
\item The 1000-class image classification scores of a deep model on the whole image are used as the contextual information for refining the 200 detection scores of each candidate bounding box. Details are given in Section \ref{Sec:Context}.
\item Average of multiple deep model outputs is used to improve the detection accuracy. Details are given in Section \ref{Sec:CombineModel}.
\item The bounding box regression in RCNN is used to reduce localization errors. 
\end{enumerate}

%
%

\subsection{Bounding box proposal by selective search}
\label{Sec:RegionSel}
Many approaches have been proposed to generate class-independent bounding box proposals. The recent approaches include objectness \cite{alexe2012measuring},  selective search \cite{Smeulders:SelectiveSearch}, category independent object proposals \cite{endres2010category}, constrained parametric min-cuts \cite{carreira2012cpmc}, combinatorial grouping \cite{arbelaez2014multiscale}, binarized normed gradients \cite{cheng2014bing}, deep learning \cite{erhan2013scalable}, and edge boxes \cite{ZitnickDollarECCV14edgeBoxes}. 
The selective search approach in \cite{Smeulders:SelectiveSearch} is adopted in order to have fair comparison with the RCNN in \cite{girshick2014rich}. We strictly followed the RCNN in using the selective search, where selective search was run in “fast mode” on each image in val$_1$, val$_2$ and test, and each image was resized to a fixed width (500 pixels) before running selective search. In this way, selective search resulted in an average of 2403 bounding box proposals per image with a 91.6\% recall of all ground-truth bounding boxes by choosing Intersection over Union (IoU) threshold as 0.5.

\subsection{Bounding box rejection}
\label{Sec:RegionRej}
On the val data, selective search generates 2403 bounding boxes per image. On average, 10.24 seconds per image are required using the Titan GPU (about 12 seconds per image using GTX670) for extracting features from bounding boxes. Features in val and test should be extracted for training SVM  or validating performance. This feature extraction takes around 2.4 days on the val dataset and around 4.7 days on the test dataset. The feature extraction procedure is time consuming and slows down the training and testing of new models. In order to speed up the feature extraction for new models, we use an existing approach, RCNN \cite{girshick2014rich} in our implementation, for rejecting bounding boxes that are most likely to be background. Denote by $\mathbf{s}_i$ the detection scores for 200 classes of the $i$th bounding box. The $i$th bounding box is rejected if the following rejection condition is satisfied:
\begin{equation}
\label{eq:Reject}
||\mathbf{s}_i||_{\infty}<T, 
\vspace{-5pt}
\end{equation}
where $||\mathbf{s}_i||_{\infty} = \max_j\{s_{i,j}\}$, $s_{i,j}$ is the $j$th element in $\mathbf{s}_i$. Since the elements in $\mathbf{s}_i$ are SVM scores, negative samples with scores smaller than $-1$ are not support vectors for SVM. When $||\mathbf{s}_i||_{\infty}<-1$, the scores are below the negative-sample margins for all the classes. We choose $T=-1.1$ as the threshold to be a bit more conservative than the margin $-1$. With the rejection condition in (\ref{eq:Reject}), $94\%$ bounding boxes are rejected and only the $6\%$ remaining windows are used for further process of DeepID-Net at the training and testing stages. The remaining $6\%$ bounding boxes result in 84.4\% recall of all ground-truth bounding boxes (at 0.5 IoU threshold), 7.2\% drop in recall compared with the 100\% bounding boxes. Since the easy examples are rejected, the DeepID-Net can focus on hard examples. 


For the remaining 6\% bounding boxes, the execution time required by feature extraction is 1.18 seconds per image on Titan GPU, about $1/9$ of the 10.24 seconds per image required for the 100\% bounding boxes. In terms of detection accuracy, bound boxing rejection can improve the mean AP by around $1\%$.

\section{Bounding box classification by DeepID-Net}
\label{Sec:DeepIdNet}

\subsection{Overview of DeepID-Net}
\label{Sec:NetOverview}
\begin{figure} 
\centering
\begin{tabular}{c}
\centerline{\epsfig{figure=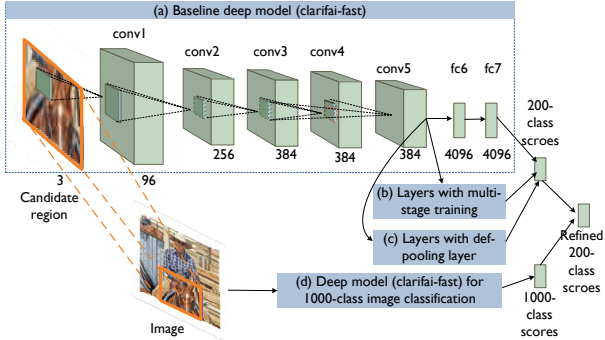,width=8cm}} \\
\end{tabular}
\caption{Overview of the DeepID-Net. It consists of four parts, (a) the baseline deep model, which is the Clarifai-fast \cite{zeiler2013visualizing} in our best-performing single model; (b) the layers with multi-stage training; (c) the layers with variable filter sizes and def-pooling layer; (d) the deep model for obtaining 1000-class image classification scores. The 1000-class image classification scores and the 200-class bounding box classification scores are combined into the refined 200-class bounding box classification scores.}
\label{Fig:DeepIDmodel}
\vspace{-10pt}
\end{figure}

An overview of the DeepID-Net is given in Fig. \ref{Fig:DeepIDmodel}. This deep model contains four parts:
\begin{enumerate}[label=(\alph*),leftmargin=12pt,noitemsep,nolistsep]
\item The baseline deep model. The input is the image region cropped by a candidate bounding box. The input image region is warped to $227 \times 227$. The Clarifai-fast in \cite{zeiler2013visualizing} is used as the baseline deep model in our best-performing single model. The Clarifai-fast model contains 5 convolutional layers (conv1-conv5) and two fully connected layers (fc6 and fc7). conv1 is the result of convolving its previous layer, the input image, with learned filters. Similarly for conv2-conv5, fc6, and fc7. Max pooling layers, which are not shown in Fig. \ref{Fig:DeepIDmodel}, are used after conv1, conv2 and conv5.
\item Fully connected layers learned by the multi-stage training scheme, which is detailed in Section \ref{Sec:NetMStage}. The input of these layers is the pooling layer after conv5 of the baseline model.
\item Layers with def-pooling layer. The input of these layers is the conv5 of the baseline model. The conv5 layer is convolved by filters with variable sizes and then the proposed def-pooling layer in Section \ref{Sec:DeformLayer} is used for learning the deformation constraint of these part filters. Parts (a)-(c) outputs the 200-class object detection scores. For the example in Fig. \ref{Fig:DeepIDmodel}, ideal output will have a high score for the object class horse but low scores for other classes for the cropped image region that contains a horse.
\item The deep model (Clarifai-fast) for obtaining the image classification scores of 1000 classes. The input is the whole image. The image classification scores are used as contextual information for refining the scores of the bounding boxes. Detail are given in Section \ref{Sec:Context}.
\end{enumerate}
Parts (a)-(d) are learned by back-propagation (BP).

\subsection{New pretraining strategy}
\label{Sec:Prtrain}
The training scheme of the RCNN in \cite{girshick2014rich} is as follows:
\begin{enumerate}[leftmargin=12pt,noitemsep,nolistsep]
\item Pretrain the deep model by using the image classification task, i.e. using image-level annotations of 1000 classes from the ImageNet Cls-Loc train data.
\item Fine-tune the deep model for the object detection task, i.e. using object-level annotations of 200 classes from the ImageNet Det train and val$_1$ data.  
\end{enumerate}
The deep model structures at the pretraining and fine-tuning stages are only different in the last fully connected layer for predicting labels ($1,000$ classes vs. $200$ classes). Except for the last fully connected layers for classification, the parameters learned at the pretraining stage are directly used as initial values for the fine-tuning stage.

The problem of the training scheme of RCNN is that image classification and object detection are different tasks, which have different requirements on the learned feature representation. 
For image classification, the whole image is used as the input and the class label of objects within the image is estimated. An object may appear in different places with different sizes in the image. Therefore, the deep model learned for image classification is required to be robust to scale change and translation of objects. 
For object detection, the image region cropped with a  tight bounding box is used as the input and the class label of objects within the bounding box is estimated. Since tight bounding box is used, robustness to scale change and translation of object  is not needed. This is the reason why bag of visual words is popular for image classification but not for detection.  The mismatch in image classification and object detection results in the mismatch in learning features for the deep model.

Another potential mismatch comes from the fact that the Cls-Loc data has $1,000$ classes, while the ImageNet detection challenge only targets on $200$ classes. However, our experimental study shows that feature representations pretrained with $1,000$ classes have better generalization capability, which leads to better detection accuracy than only selecting the $200$ classes from the Cls-Loc data for pretraining. 

 Since the ImageNet Cls-Loc data provides object-level bounding boxes for 1000 classes, which is more diverse in content than the ImageNet Det data with 200 classes, we use the images regions cropped by these bounding boxes as the training samples to pretain the baseline deep model. We propose two new pretraining strategies that bridge the image- vs. object-level annotation gap in RCNN. 

Scheme 1 is as follows:
\begin{enumerate}[leftmargin=12pt,noitemsep,nolistsep]
\item Pretrain the deep model by using image-level annotations of $1,000$ classes from the ImageNet Cls-Loc train data.
\item Fine-tune the deep model with object-level annotations of $1,000$ classes from the ImageNet Cls-Loc train data. The parameters trained from Step (1) is used as initialization.
\item Fine-tune the deep model for the second time by using object-level annotations of $200$ classes from the ImageNet Det train and val$_1$ data.  The parameters trained from Step (2) are used as initialization.
\end{enumerate}
Scheme 1 uses pretraining on 1000-class object-level annotations as the intermediate step to bridge the gap between 1000-class image classification task and 200-class object detection task.

Scheme 2 is as follows:
\begin{enumerate}[leftmargin=12pt,noitemsep,nolistsep]
\item Pretrain the deep model with object-level annotations of $1,000$ classes from the ImageNet Cls-Loc train data.
\item Fine-tune the deep model for the 200-class object detection task, i.e. using object-level annotations of 200 classes from the ImageNet Det train and val$_1$ data.  Use the parameters in Step (1) as initialization.
\end{enumerate}
Scheme 2 removes pretraining on the image classification task and directly uses object-level annotations to pretrain the deep model.
Compared with the training scheme of RCNN, experimental results on ImageNet Det val$_2$ found that scheme 1 improves mean AP by 1.6\% and scheme 2 improves mean AP by 4.4\%.

The baseline deep model is pretrained using the approach discussed above. The layers with mulit-stage training and def-pooling layers in Fig. \ref{Fig:DeepIDmodel} are randomly initialized and trained at the fine-tuning stage. 

\subsection{Fully connected layers with multi-stage training}
\label{Sec:NetMStage}

\emph{Motivation.}
Multi-stage classifiers have been widely used in object detection and achieved great success. With a cascaded structure, each classifier processes a different subset of data  \cite{Viola:FaceDetect, Dollar:ChnFtrs, bourdev2005robust,Felzenszwalb:cascade10, Vedaldi:MultKernel}. However, these classifiers are usually trained sequentially without joint optimization. In this paper, we propose a new deep architecture that can jointly train multiple classifiers through several stages of back-propagation. 
Each stage handles samples at a different difficulty level. 
Specifically the first stage of deep CNN handles easy samples, the second stage of deep model processes more difficult samples which cannot be handled in the first stage, and so on. 
Through a specific design of the training strategy, this deep architecture is able to simulate the cascaded classifiers by mining hard samples to train the network stage-by-stage. 
Our recent work \cite{Xingyu2013DeepPed} has explored the idea of multi-stage deep learning, but it was only applied to pedestrian detection. In this paper, we apply it to general object detection on ImageNet. 

\begin{figure} 
\centering
\centerline{\epsfig{figure=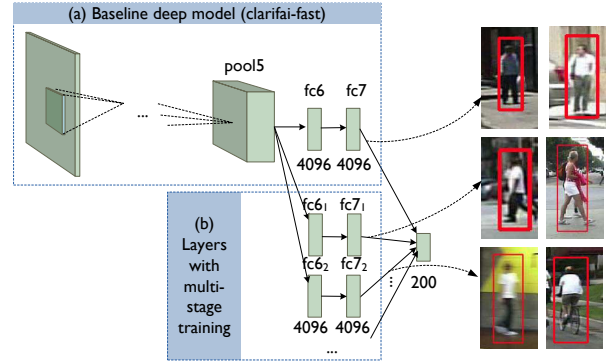,width=8cm}}
\caption{The baseline deep model and fully connected layers with multi-stage training. The layer pool5 is result of max pooling over the conv5 layer in Fig. \ref{Fig:DeepIDmodel}. Different stages of classifiers deal with samples of different difficulty levels.}
\label{Fig:ModelMS}
\vspace{-10pt}
\end{figure}

\emph{Denotations.}
The pooling layer after conv5 is denoted by pool5.
As shown in Fig. \ref{Fig:ModelMS}, besides fc6, pool5 is connected to $T$ extra fully connected layers of sizes 4096. Denote the $T$ extra layers connected the pool5 layer as fc6$_1$, fc6$_2$, $\cdots$, fc6$_T$. Denote fc7$_1$, fc7$_2$, $\cdots$, fc7$_T$ as the $T$ layers separately connected to the layers fc6$_1$, fc6$_2$, $\cdots$, fc6$_T$. Denote the weight connected to fc$l$$_T$ by $\mathbf{W}_{l,t}$, $l=6,7, t=1, \cdots, T$. Denote the weights from fc7$_t$ to classification scores as $\mathbf{W}_{8,t}$, $t = 1, \cdots, T$. The path from pool5, fc6$_t$,  fc7$_t$ to classification scores can be considered as the extra classifier at stage $t$.

The multi-stage training procedure is summarized in Algorithm \ref{alg:training}.
It consists of two steps. 
\begin{itemize}
	\item Step 1 (2 in Algorithm \ref{alg:training}): BP is used for fine-tuning all the parameters in the baseline deep model.
  \item Step 2.1 (4 in Algorithm \ref{alg:training}): parameters $\mathbf{W}_{l,t}, t=6, 7$ are randomly initialized at stage $t$ in order to search for extra discriminative information in the next step.
  \item Step 2.2 (5-6 in Algorithm \ref{alg:training}): multi-stage classifiers $\mathbf{W}_{l,t}$ for $l=6,7, t=1, \cdots, T$ are trained using BP stage-by-stage. In stage $t$, classifiers $\mathbf{W}_{l,t}$ up to $t$ are jointly updated.
\end{itemize}
The baseline deep model is first trained by excluding extra classifiers to reach a good initialization point. Training this simplified model avoids overfitting. Then the extra classifiers are added stage-by-stage. At stage $t$, all the existing classifiers up to layer $t$ are jointly optimized. Each round of optimization finds a better local minimum around the good initialization point reached in the previous training stages.



\begin{algorithm}[t]
\KwIn{Training set: Warped images and their labels from the fine-tuning training data\\
~~~~~~~~~~~~Parameters $\Theta$ for the baseline deep model \\
~~~~~~~~~~~~obtained by pretraining.}
\KwOut{Parameters $\Theta$ for the baseline deep model, \\
~~~~~~~~~~~~~~~Parameters $\mathbf{W}_{l,t}$, $l=6,7,8, t=1, \cdots, T$ for \\
~~~~~~~~~~~~~~~the extra layers. }

Set elements in $\mathbf{W}_{l,t}$ to be 0\;
 BP to fine-tune $\Theta$, while keeping $\mathbf{W}_{l,t}$ as 0\;
\For{$t$=1 to $T$}
{
Randomly initialize $\mathbf{W}_{l,t}$, $l=6, 7$\;
Use BP to update parameters $\mathbf{W}_{l,t}$, $l=6, 7, 8$ while fixing $\Theta$ and $\mathbf{W}_{l,1}, \cdots, \mathbf{W}_{l,t-1}$\;
Use BP to update parameters $\Theta$ and $\mathbf{W}_{l,1}, \cdots, \mathbf{W}_{l,t}$, $l=6, 7, 8$\;
}
Output $\Theta$ and $\mathbf{W}_{l,t}$, $l=6, 7, 8, t=1, \cdots, T$.\
\caption{Stage-by-Stage Training}
\label{alg:training}
\end{algorithm}

In the stage-by-stage training procedure, classifiers at the previous stages jointly work with the classifier at the current stage in dealing with misclassified samples. Existing cascaded classifiers only pass a single score to the next stage, while our deep model uses multiple hidden nodes to transfer information. 

Detailed analysis on the multi-stage training scheme is provided in \cite{Xingyu2013DeepPed}. A brief summary is given as follows:
First, it simulates the soft-cascade structure. A new classifier is introduced at each stage to help deal with misclassified samples while the correctly classified samples have no influence on the new classifier. Second, the cascaded classifiers are jointly optimized at stage $t$ in step 2.2, such that these classifiers can better cooperate with each other. Third, the whole training procedure helps to avoid overfitting. The supervised stage-by-stage training can be considered as adding regularization constraints to parameters, i.e. some parameters are constrained to be zeros in the early training strategies. At each stage, the whole network is initialized with a good point reached by previous training strategies and the additional classifiers deal with misclassified samples. It is important to set $\mathbf{W}_{l,t}=0$ in the previous training strategies; otherwise, it become standard BP. With standard BP, even an easy training sample can influence any classifier. Training samples will not be assigned to different classifiers according to their difficulty levels. The parameter space of the whole model is huge and it is easy to overfit.

\begin{figure} 
\centering
\centerline{\epsfig{figure=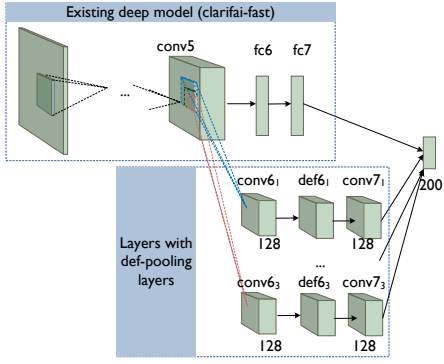,width=6cm}}
\caption{The baseline deep model and def-pooling layers.}
\label{Fig:ModelDef}
\vspace{-10pt}
\end{figure}

\subsection{The def-pooling layer}
\subsubsection{Generating the part detection map} \label{ssec:partmap}
Since object parts have different sizes, we design filters with variable sizes and convolve them with the conv5 layer in the baseline model. Fig. \ref{Fig:ModelDef} shows the layers with def-pooling layers.  It contains the following four parts:
\begin{enumerate}[label=(\alph*),leftmargin=12pt,noitemsep,nolistsep]
\item The conv5 layer is convolved by filters of sizes $3\times 3$, $5\times 5$, and $9\times 9$ separately in order to obtain the \emph{part detection maps} of 128 channels, which are denoted by conv6$_1$, conv6$_2$, and conv6$_3$ as shown in Fig. \ref{Fig:ModelDef}. In comparison, the path from conv5, fc6, fc7 to classification score can be considered as a holistic model.

\item Part detection maps are separately fed into the \emph{def-pooling layers} denoted by def6$_1$, def6$_2$, and def6$_3$ in order to learn their deformation constraints. 

\item The output of def-pooling layers, i.e. def$6_1$, def$6_2$, and def$6_3$, are separately convolved with filters of sizes $1\times 1$ with 128 channels to produce outputs conv7$_1$, conv7$_2$, and conv7$_3$, which can be considered as fully connected layers over the 128 channels for each location. 

\item The fc7 in the Clarifai-fast and the output of layers conv7$_1$, conv7$_2$, and conv7$_3$ are used for estimating the class label of the candidate bounding box.
\end{enumerate}



\subsubsection{Learning the deformation}
\label{Sec:DeformLayer}
\emph{Motivation.} The effectiveness of learning deformation constraints of object parts has been proved in object detection by many existing non-deep-learning detectors, e.g. \cite{LatSVMObj}. However, it is missed in current deep learning models. In deep CNN models, max pooling and average pooling are useful in handling deformation but cannot learn the deformation constraint and geometric model of object parts. We design the def-pooling layer for deep models so that the deformation constraint of object parts can be learned by deep models.

Denote $\mathbf{M}$ of size $V\times H$ as the result of the convolutional layer, e.g. conv6$_1$.
The def-pooling layer takes small blocks of size $(2R+1)\times (2R+1)$ from the $\mathbf{M}$ and subsamples $\mathbf{M}$ to $\mathbf{B}$ of size $\frac{V}{k_x}\times \frac{H}{k_y}$ to produce single output from each block as follows: 
{\small
\vspace{-5pt}
\begin{equation}
\label{eq:GenDefMap2}
b^{(x,y)} = \max_{i,j\in\{-R, \cdots, R\}}{\{m^{(k_x\cdot x+i,k_y \cdot y+j)} - \sum_{n=1}^N c_{n}d^{i,j}_{n}\}}, 
\vspace{-5pt}
\end{equation}}
where $(k_x\cdot x,k_y \cdot y)$ is the center of the block, $k_x$ and $k_y$ are subsmpling steps, $b^{(x,y)}$ is the $(x,y)$th element of $\mathbf{B}$. $c_{n}$ and $d^{i,j}_{n}$ are deformation parameters to be learned. 

\emph{Example 1}. Suppose $c_{n}=0$, then there is no penalty for placing a part with center $(k_x\cdot x,k_y \cdot y)$ to any location in $\{(k_x\cdot x+i,k_y \cdot y+j)| i,j=-R, \ldots R\}$. In this case, the def-pooling layer degenerates to max-pooling layer with subsampling step $(k_x, k_y)$ and kernel size $(2R+1)\times (2R+1)$. Therefore, the difference between def-pooling and max-pooling is the term $- \sum_{n=1}^N c_{n}d^{i,j}_{n}$ in (\ref{eq:GenDefMap2}), which is the deformation constraint learned by def-pooling. In short, def-pooling is max-pooling with deformation constraint.

\emph{Example 2}. Suppose $V=k_y$, $H=k_x$, $i=1, \cdots, V$, and $j=1, \cdots, H$, then the def-pooling layer degenerates to the deformation layer in \cite{Ouyang2013JointDeep}. There is only one output for $\mathbf{M}$ in this case. The deformation layer can represent the widely used quadratic deformation constraint in the deformable part-based model \cite{LatSVMObj}. Details are given in Appendix A. Fig. \ref{Fig:DefLayer} illustrates this example.

\begin{figure} 
\centering
\centerline{\epsfig{figure=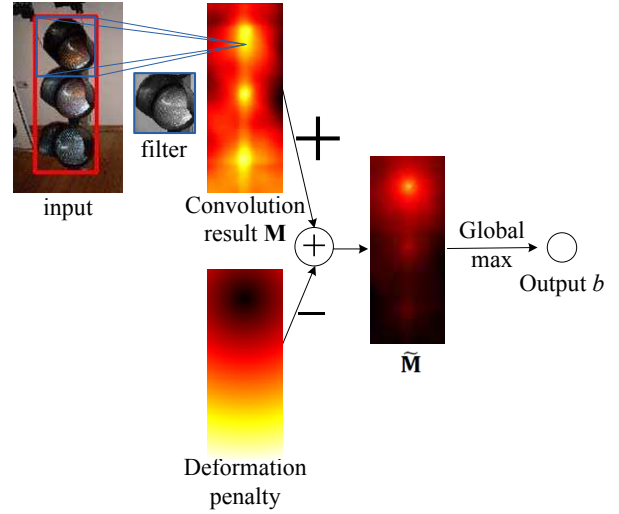,width=8cm}}
\caption{The deformation layer when deformation map is defined in (\ref{eq:DefMap5}). Part detection map $\mathbf{M}$ and deformation constraint are summed up to obtain the summed map $\tilde{\mathbf{M}}$. Global max pooling is then performed on $\tilde{\mathbf{M}}$ to obtain the score $b$.}
\label{Fig:DefLayer}
\end{figure}

\emph{Example 3}. Suppose $N=1$ and $c_n=1$, then the deformation constraint $d^{i,j}_{1}$ is learned for each displacement bin $(i,j)$ from the center location $(k_x\cdot x,k_y\cdot y)$. In this case, $d^{i,j}_{1}$ is the deformation cost of moving an object part from the center location $(k_x\cdot x,k_y\cdot y)$ to location $(k_x\cdot x+i,k_y \cdot y+j)$. As an example, if $d^{0,0}_{1}=0$ and $d^{i,j}_{1}=\infty$ for $(i,j)\neq (0,0)$, then the part is not allowed to move from the center location $(k_x\cdot x,k_y\cdot y)$ to anywhere. As the second example, if $d^{i,j}_{1}=0$ for $j<=0$ and $d^{i,j}_{1}=\infty$ for $j>0$, then the part can move freely upward but should not move downward. As the third example, if $d^{0,0}_{1}=0$  and $d^{i,j}_{1}=1$ for $(i,j)\neq (0,0)$, then the part has no penalty at the center location $(k_x\cdot x,k_y\cdot y)$ but has penalty 1 elsewhere. The $R$ in controls the movement range. Objects are only allowed to move within the horizontal and vertical range $[-R\ R]$  from the center location.

\begin{figure} 
\centering
\centerline{\epsfig{figure=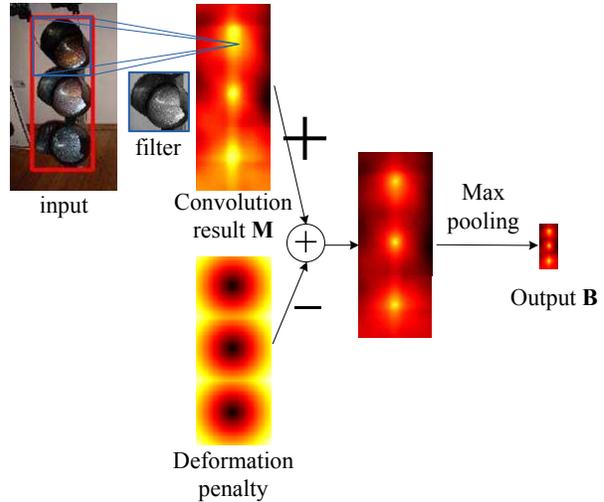,width=8cm}}
\caption{The def-pooling layer. Part detection map and deformation constraint are summed up. Block-wise max pooling is then performed on the summed map to obtain the output $\mathbf{B}$ of size $\frac{H}{k_y}\times \frac{V}{k_x}$.}
\label{Fig:DefLayerGen}
\end{figure}

\begin{figure} 
\centering
\centerline{\epsfig{figure=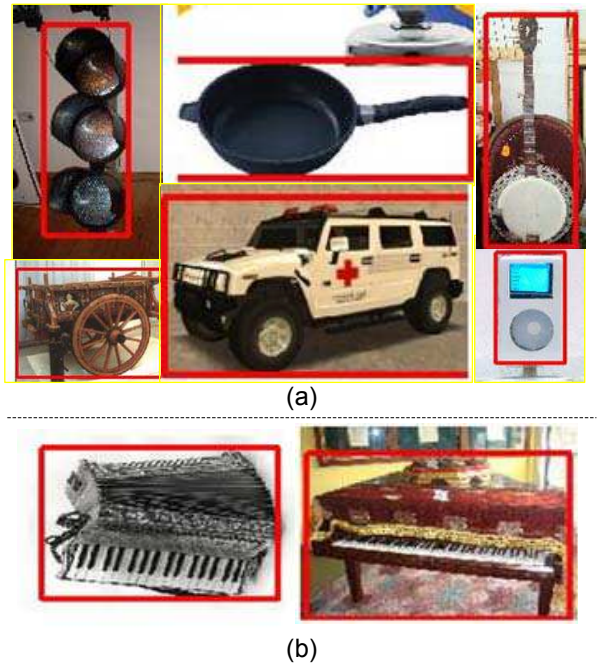,width=8cm}}
\caption{The circular patterns (a) and musical instrument key patterns (b) shared across different object classes.}
\label{Fig:Share_pattern}
\end{figure}

The deformation layer was proposed in our recently published work \cite{Ouyang2013JointDeep}, which showed significant improvement in pedestrian detection.
The def-pooling layer in this paper is different from the deformation layer in \cite{Ouyang2013JointDeep} in the following aspects. 
\begin{enumerate}[leftmargin=12pt,noitemsep,nolistsep]
\item The work in \cite{Ouyang2013JointDeep} only allows for one output, while this paper is block-wise pooling and allows for multiple output at different spatial locations. Because of this difference, the deformation layer can only be put after the final convolutional layer, while the def-pooling layer can be put after any convolutional layer like the max-pooling layer. Therefore, the def-pooling layer can capture geometric deformation at all the levels of abstraction, while the deformation layer was only applied to a single layer corresponding to pedestrian body parts. 

\item It was assumed in \cite{Ouyang2013JointDeep} that a pedestrian only has one instance of a body part, so each part filter only has one optimal response in a detection window. In this work, it is assumed that an object has multiple instances of its part (e.g. a building has many windows, a traffic light has many light bulbs), so each part filter is allowed to have multiple response peaks. This new model is more suitable for general object detection. For example, the traffic light can have three response peaks to the light bulb in Fig. \ref{Fig:DefLayerGen} for the def-pooling layer but only one peak in Fig. \ref{Fig:DefLayer} for the deformation layer in \cite{Ouyang2013JointDeep}.

\item The approach in \cite{Ouyang2013JointDeep} only considers one object class, e.g. pedestrians. In this work, we consider 200 object classes. The patterns can be shared across different object classes. As shown in Fig. \ref{Fig:Share_pattern}, circular patterns are shared in wheels for cars, light bulb for traffic lights, wheels for carts and keys for ipods. Similarly, the pattern of instrument keys is shared in accordion and piano. In this work, our design of the deep model in Fig. \ref{Fig:DefLayerGen} considers this property and learns the shared patterns through the layers conv6$_1$, conv6$_2$ and conv6$_3$ and use these shared patterns for 200 object classes.
\end{enumerate}

\subsection{Fine-tuning the deep model with hinge-loss}
RCNN fine-tunes the deep model with softmax loss, then fixes the deep model and uses the hidden layers fc7 as features to learn 200 one-versus-all SVM classifiers. This scheme results in extra time required for extracting features from training data. With the bounding box rejection, it still takes around 60 hours to prepare features from the ILSVRC2013 Det train and val$_1$ for SVM training. In our approach, we replace the softmax loss of the deep model by hinge loss when fine-tuning deep models. The deep model fine-tuning and SVM learning steps in RCNN are merged into one step in our approach. In this way, the extra training time required for extracting features is saved in our approach. 

\subsection{Sub-box features}
\begin{figure} 
\centering
\centerline{\epsfig{figure=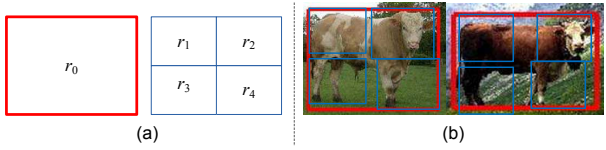,width=8cm}}
\caption{A box $r_0$ with its four sub-boxes $r_1, \cdots, r_4$(a) and examples for the bounding boxes on cattle (b).}
\label{Fig:Subregion}
\end{figure}

A bounding box denoted by $r_0$ can be divided into $N$ sub-boxes $r_1, \cdots, r_N$, $N=4$ in our implementation. $r_0$ is called the root box in this paper. For example, the bounding box for cattle in Fig. \ref{Fig:Subregion} can be divided into 4 sub-boxes corresponding to head, torso, forelegs and hind legs. The features of these sub-boxes can be used to improve the object detection accuracy. In our implementation, sub-boxes have half the width and height of the root box $r_0$. The four sub-boxes locate at the four corners of the root box $r_0$. Denote $\mathbf{B}_s$ as the set of bounding boxes generated by selective search. The features for these bounding boxes have been generated by deep model. 
The following steps are used for obtaining the sub-box features:
\begin{enumerate}[leftmargin=12pt,noitemsep,nolistsep]
\item For a sub-box $r_n$, $n=1, \cdots, 4$, its overlap with the the boxes in $\mathbf{B}_s$ is calculated. The box in $\mathbf{B}_s$ having the largest IoU with $r_n$ is used as the selected box $\mathbf{b}_{s,n}$ for the sub-box $r_n$. 
\item The features of the selected box $\mathbf{b}_{s,n}$ are used as the features $\mathbf{f}_n$ for sub-box $r_n$.
\item Element-wise max-pooling over the four feature vectors $\mathbf{f}_n$ for $n=1, 2, 3, 4$ is used for obtaining max-pooling feature vector $\mathbf{f}_{max}$, i.e. $f_{i, max} = \max_{n=1}^{4}{f_{i,n}}$, where $f_{i, max}$ is the $i$th element in $\mathbf{f}_{max}$ and $f_{i,n}$ is the $i$th element in $\mathbf{f}_{n}$ .
\item Element-wise average-pooling over the four feature vectors $\mathbf{f}_n$ for $n=1, 2, 3, 4$ is used for obtaining average-pooling feature vector $\mathbf{f}_{avg}$, i.e. $f_{i, avg} = \frac{1}{4}\sum_{n=1}^{4}{f_{i,n}}$, where $f_{i, avg}$ is the $i$th element in $\mathbf{f}_{avg}$.
\item Denote the feature for the root box as $\mathbf{f}_0$. $\mathbf{f}_0$, $\mathbf{f}_{max}$, and $\mathbf{f}_{avg}$ are concatenated as the combined feature $\mathbf{f}=\{\mathbf{f}_0, \mathbf{f}_{max}, \mathbf{f}_{avg}\}$. 
\item $\mathbf{f}$ is used as the feature for box $r_0$. Linear SVM is used as the object detection classifier for these features.
\end{enumerate}
The hierarchical structure of selective search has provided us with the opportunity of reusing the features computed for small root box as the sub-box for large root box. The sub-box features need not be computed and is directly copied from the features computed for bounding boxes of selective search. In this way, the execution time for computing features is saved for sub-boxs. Another good property is that the selected bounding boxes for sub-boxes are allowed to move, which improves the robustness to the translation of object parts. With sub-box features, the mAP improves by 0.5\%.

\subsection{Contextual modeling}
\label{Sec:Context}
The model learned for the image classification task takes the scene information into consideration while the model for object detection focuses on local boxes. Therefore, the image classification scores provides contextual information for object detection. We use 1000-class image classification scores as the contextual features. The steps of using contextual modeling is as follows:
\begin{enumerate}[leftmargin=12pt,noitemsep,nolistsep]
\item The 1000-class scores of image classification and 200 scores of object detection are concatenated as the 1200 dimensional feature vector. 
\item Based on the 1200 features, 200 one-versus-all linear SVM classifiers are learned for 200 object detection classes. At the testing stage, the classification scores obtained by linear weighting of the 1200 dimensional features are used as the refined score for each candidate bounding box.
\end{enumerate} 
For the object detection class volleyball, Fig. \ref{Fig:Context} shows the weights for the 1000 image classes. It can be seen that image classes bathing cap and golf ball suppress the existence of volleyball with negative weight while the image class volleyball enhances the existence of detection class volleyball. The bathing cap often appears near the beach or swimming pool, where it is unlikely to have volleyball. 

\begin{figure} 
\centering
\centerline{\epsfig{figure=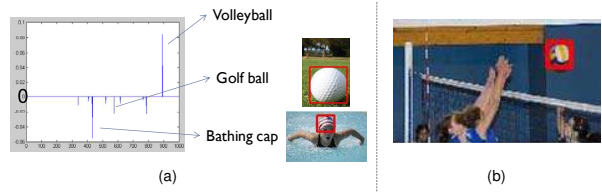,width=8cm}}
\caption{The weights of image classification scores (a) for the object detection class volleyball (b). }
\label{Fig:Context}
\end{figure}

 \section{Combining models with high diversity}
\label{Sec:CombineModel}
In existing model combination approaches \cite{zeiler2013visualizing, Krizhevsky:ImageNetCNN, he2014spatial}, the same deep architecture is used. Models are different in spatial locations or learned parameters. In our model averaging scheme, we learn models under several settings. The settings of the 10 models we used for model averaging when submitted to ILSVRC2014 challenge are shown in Table \ref{Table:ModelSel}. The 10 models are different in net structure, pretraining scheme, loss functions for the deep model training, adding def-pooling layer/multi-stage training/sub-box features or not, and whether to do bounding box rejection or not. In our current implementation, the def-pooling layers, multi-stage training and sub-box features are added to different deep models separately without being integrated together, although such integration can be done in the future work. Models generated in this way have high diversity and are complementary to each other in improving the detection results. The 10 models were selected with greedy search based on performance on val$_2$. The mean AP (mAP) of averaging these 10 models is $40.9\%$ on val$_2$, and its mAP on the test data of ILSVRC2014 is 40.7\%, ranking \#2 in the challenge. 
After the deadline of ILSVRC2014, our deep models were further improved. Running model averaging again, the selected models and their configurations are shown in Table \ref{Table:ModelAvg}. The mAP on val$_2$ is 42.4\%. 

In existing works and the model averaging approach described above, the same model combination is applied to all the $200$ classes in detection. However, we observe that the effectiveness of different models varies a lot across different object categories. Therefore, it is better to do model selection for each class separately. With this strategy, we achieve mAP $45\%$ on val$_2$. 

\begin{table}
\setlength{\tabcolsep}{1.5pt}
\centering
\caption{Models used for model averaging submitted to ILSVRC2014. The result of mAP is on val$_2$. For net design, A denotes AlexNet, C denotes Clarifai-fast, D-D denotes DeepID-Net with def-pooling layers, D-MS denotes DeepID-Net with multi-stage training. In A and C, only the baseline deep model (Clarifai-fast or AlexNet) is used without def-pooling layers or multi-stage training. In D-D and S-MS, the baseline deep model is chosen as Clarifai-fast, and extra layers from def-pooling or multi-stage training are included. For pretrain,  \cite{girshick2014rich} denotes the pretraining scheme of RCNN, 1 denotes the Scheme 1 in Section \ref{Sec:Prtrain}, 2 denotes the Scheme 2 in Section \ref{Sec:Prtrain}. }
{\small
\begin{tabular}{ccccccccccc}
\hline	 model number  & 1 & 2 & 3 & 4 & 5 & 6 & 7 & 8 & 9 & 10 \\
         bbox rejection & y &n  & y & y & y & y & y & y & y & y\\
         net design  & A & A& C&C&D-D&D-D&D-MS&D-D &D-D &D-D \\
				 Pretrain   & \cite{girshick2014rich} & 1& \cite{girshick2014rich}&1&1&1&2&2&2&2 \\
				 loss of net & s & s& s&h&h&h&h&h&h&h \\
\hline   mAP ($\%$)         & 31.0	& 31.2	& 32.1	& 33.6	& 35.3	& 36.0	& 37.0	& 37.0	& 37.1	& 37.4 \\
\hline
\end{tabular}
}
\label{Table:ModelSel}
\vspace{-10pt}
\end{table}

\begin{table*}
\setlength{\tabcolsep}{1.5pt}
\centering
\caption{Experimental results on ILSVRC2014  for top ranked approaches. }
{\small
\begin{tabular}{*6c|cc}
\hline	 
	     approach  & RCNN\cite{girshick2014rich}  &Berkeley Vision&UvA-Euvision	& DeepInsight&GoogLeNet &ours& ours new\\
\hline   mAP ($\%$) on val$_2$    & 31.0 &33.4&n/a & n/a&44.5  &40.9 & 45\\
\hline   mAP ($\%$) on test       & 31.4  &34.5&35.4& 40.5 &43.9  &40.7& n/a\\
\hline
\end{tabular}
}
\label{Table:allMethod}
\vspace{-10pt}
\end{table*}

\begin{table}
\setlength{\tabcolsep}{1.5pt}
\centering
\caption{Experimental results for model averaging on ILSVRC 2014. Fore averaging scheme, all-cls denotes the greedy search in which all classes share the same set of models for averaging, per-cls denotes the greedy search in which different classes have different model combinations. Since our results got improved after the competition deadline, both results submitted before and after the deadline are reported on both val$_2$ and test data.}
{\small
\begin{tabular}{ccccccccccc}
\hline	 Averaging scheme  & all-cls & all-cls & all-cls& per-cls \\
         After deadline & n &n  & y & y \\
         evaluation data  & val$_2$ &test& val$_2$ & val$_2$ \\
\hline   mAP ($\%$)         & 40.9	&40.7	& 42.4	& 45		 \\
\hline
\end{tabular}
}
\label{Table:ModelAvg}
\vspace{-10pt}
\end{table}
%

\section{Experimental Results}
The ImageNet Det val$_2$ data is used for evaluating separate components and the ImageNet Det test data is used for evaluating the overall performance.
The RCNN approach in \cite{girshick2014rich} is used as the baseline for comparison. The source code provided by the authors are used for repeating their results. Without bounding box regression, we obtain mean AP 29.9 on val$_2$, which is close to the 29.7 reported in \cite{girshick2014rich}. Table \ref{Table:allMethod} summarizes the results from ILSVRC2014 object detection challenge. It includes the best results on test data submitted to ILSVRC2014 from our team, GoogleNet, DeepInsignt, UvA-Euvision, and Berkeley Vision, which ranked top five among all the teams participating in the challenge. It also includes our most recent results on test data obtained after the competition deadline. All these best results were obtained with model averaging.

\begin{table}
\setlength{\tabcolsep}{1.5pt}
\centering
\caption{Ablation study of bounding box (bbox) rejection and baseline deep model on ILSVRC2014 val$_2$. }
{\small
\begin{tabular}{*4c}
\hline	 bbox rejection? & n & y& y\\
				 deep model  & A-net & A-net& C-net \\
\hline   mAP ($\%$)        & 29.9  & 30.9 & 31.8  \\
         meadian AP ($\%$)  & 28.9  & 29.4 & 30.5  \\
\hline
\end{tabular}
}
\label{Table:RejAndModel}
\vspace{-10pt}
\end{table}

\subsection{Ablation study}
\subsubsection{Investigation on bounding box rejection and baseline deep model}
As shown in Fig. \ref{Fig:DeepIDmodel}, a baseline deep model is used in our DeepID-Net. The baseline deep model using the AlexNet in \cite{Krizhevsky:ImageNetCNN} is denoted as A-net and the baseline deep model using the clarifai-fast in \cite{zeiler2013visualizing} is denoted as C-net. 
Table \ref{Table:RejAndModel} shows the results for different baseline deep model and bounding box rejection choice. Except for the two components investigated in Table \ref{Table:RejAndModel}, other components are the same as RCNN, while the new training schemes and new components introduced in Section \ref{Sec:DeepIdNet} are not included. The baseline is RCNN, the first column in Table \ref{Table:RejAndModel}.
Based on the RCNN approach, applying bounding box rejection improves mAP by 1\%. Therefore, bounding box rejection not only saves the time for training and testing new models but also improves detection accuracy.  Based on the bounding box rejection step, Clarifai-fast \cite{zeiler2013visualizing} performs better than AlexNet in \cite{Krizhevsky:ImageNetCNN}, with 0.9\% mAP improvement. 

\subsubsection{Investigation on different pretraining schemes}
There are two different sets of data used for pretraining the baseline deep model. The ImageNet Cls train data with 1000 classes and the ImageNet Det train and val$_1$ data with 200 classes. There are two different annotation levels, image and object. Investigation on the combination of image class number and annotation levels is shown in Table \ref{Table:Pretrain}. When producing these results, other new components introduced in Section 5.3-5.7 are not included. Using image-level annotation, pretraining on 1000 classes performs better than pretraining on 200 classes by 9.2\% mAP. Using the same 1000 classes, pretraining on object-level-annotation peforms better than pretraining on image-level annotation by 4.4\% mAP for A-net and 4.2\% for C-net.
This experiment shows that object-level annotation is better than image-level annotation in pretraining deep model. Pretraining with more classes improves the generalization capability of the learned feature representations. 

There are two schemes in using the ImageNet object-level annotations of 1000 classes in Section \ref{Sec:Prtrain}. Scheme 1 pretrains on the image-level 1000-class annotation, first fine-tunes on object-level 1000-class annotation, and then fine-tunes again on object-level 200-class annotations. Scheme 2 does not pretrain on the image-level 1000-class annotation and directly pretrains on object-level 1000-class annotation. 
As shown in Table \ref{Table:PretrainScheme}, Scheme 2 performs better than Scheme 1 by 2.6\% mAP. 
This experiment shows that image-level annotation is not needed in pretraining deep model when object-level annotation is available. 

\begin{table}
\setlength{\tabcolsep}{1.5pt}
\centering
\caption{Ablation study of pretraining datasets and net structures on ILSVRC2014 val$_2$.}
{\small
\begin{tabular}{*6c}
\hline   net structure  	& A-net 	& A-net	& A-net	& C-net & C-net  \\
				 bbox rejection & n 			& n			& n			& y  & y\\
			   class number 		& 200 		& 1000 	& 1000	& 1000 & 1000\\
			   annotation level & image 	& image & object 	& image& object\\
\hline   mAP ($\%$)  						& 20.7 		&29.9 	& 34.3	& 31.8 & 36.0 \\
			   meadian AP ($\%$) 			&17.8			& 28.9 	& 34.9	& 30.5 & 34.9  \\
\hline
\end{tabular}
}
\label{Table:Pretrain}
\vspace{-10pt}
\end{table}

\begin{table}
\setlength{\tabcolsep}{1.5pt}
\centering
\caption{Ablation study of the two pretraining schemes in Section \ref{Sec:Prtrain} on ILSVRC2014 val$_2$. Scheme 1 uses the image-level annotation while scheme 2 does not. }
{\small
\begin{tabular}{*6c}
\hline   net structure  	& A-net & A-net & C-net & C-net  \\
				 bbox rejection & n  & n& y  & y\\
				 pretraining scheme & 1  & 2& 1  & 2\\
\hline   mAP ($\%$) 						& 31.2 		& 34.3 & 33.4 		& 36.0 \\
			   meadian AP ($\%$) 			& 29.7 		& 33.4 & 33.1			& 34.9  \\
\hline
\end{tabular}
}
\label{Table:PretrainScheme}
\vspace{-10pt}
\end{table}

\subsubsection{Investigation on deep model designs}
Based on the pretraining scheme 2 in Section \ref{Sec:Prtrain}, different deep model structures are investigated and results are shown in Table \ref{Table:designs}. Our DeepID-Net that uses multi-stage training for multiple fully connected layers in Fig. \ref{Fig:ModelMS} is denoted as D-MS.  Our DeepID-Net that uses def-pooling layers as shown in Fig. \ref{Fig:ModelDef} is denoted as D-Def. 
Using the C-net as baseline deep moel, the DeepID-Net that uses multi-stage training in Fig. \ref{Fig:ModelMS} improves mAP by 1.5\%. 
Using the C-net as baseline deep moel, the DeepID-Net that uses def-pooling layer in Fig. \ref{Fig:ModelDef} improves mAP by 2.5\%. This experiment shows the effectiveness of the multi-stage training and def-pooling layer for generic object detection.

\begin{table}
\setlength{\tabcolsep}{1.5pt}
\centering
\caption{Ablation study of the different net structures on ILSVRC2014 val$_2$. }
{\small
\begin{tabular}{*5c}
\hline   net structure  	& A-net& C-net & D-MS & D-Def  \\
				 bbox rejection & n  & y& y  & y \\
				 pretraining scheme &2 & 2& 2  & 2 \\
\hline   mAP ($\%$) 						& 34.3  & 36.0 & 37.5 & 38.5\\
			   meadian AP ($\%$) 			&  33.4 & 34.9 & 36.4 & 37.4\\
\hline
\end{tabular}
}
\label{Table:designs}
\vspace{-10pt}
\end{table}

\subsubsection{Investigation on the overall pipeline}
Table \ref{Table:overall} and Table \ref{Table:overall2} summarize how performance gets improved by adding each component step-by-step into our pipeline. RCNN has mAP $29.9\%$. With bounding box rejection, mAP is improved by about $1\%$, denoted by $\sim 1\%$. Based on that, changing A-net to C-net improves mAP by $\sim 1\%$. Replacing image-level annotation by object-level annotation for pretraining, mAP increases by $\sim 4\%$. The def-pooling layer further improves mAP by $2.5\%$. After adding the contextual information from image classification scores, mAP increases by $\sim 1\%$. Bounding box regression improves mAP by $\sim 1\%$. With model averaging, the best result is $45\%$. Table \ref{Table:overall2} summarizes the contributions of difference components.  More results on the test data will be available in the next version soon.

\begin{table*}
\setlength{\tabcolsep}{1.5pt}
\centering
\caption{Ablation study of the overall pipeline for single model tested on ILSVRC2014 val2. It shows the mean AP after adding each key component step-by-step.}
{\small
\begin{tabular}{*8cc}
\hline   detection pipeline  	& RCNN& +bbox     & A-net & image to bbox  & +Def  & +context & +bbox  &  \\
															&     & rejection & to C-net & pretrain &  pooling &  & regression & \\
\hline   mAP ($\%$) 					    	&29.9& 30.9& 31.8& 36.0 & 38.5 &39.2&40.1&\\
			   meadian AP ($\%$) 			    &28.9& 29.4& 30.5& 34.9 & 37.4 &38.7&40.3&\\
\hline
\end{tabular}
}
\label{Table:overall}
\vspace{-10pt}
\end{table*}

\begin{table*}
\setlength{\tabcolsep}{1.5pt}
\centering
\caption{Ablation study of the overall pipeline for single model tested on ILSVRC2014 val2. It summarizes the contributions from each key components. }
{\small
\begin{tabular}{*8c|c}
\hline   detection pipeline  	& RCNN& +bbox     & A-net & image to bbox  & +Def  & +context & +bbox  & model  \\
															&     & rejection & to C-net & pretrain &  pooling &  & regression & averaging \\
\hline   mAP ($\%$) 					    	&29.9& +$\sim 1\%$& +$\sim 1\%$& +$\sim 4\%$ & +2.5\% &+$\sim 1\%$&+$\sim 1\%$&$45\%$\\
\hline
\end{tabular}
}
\label{Table:overall2}
\vspace{-10pt}
\end{table*}

\section{Appedix A: Relationship between the deformation layer and the DPM in \cite{LatSVMObj}}
The quadratic deformation constraint in \cite{LatSVMObj} can be represented as follows:
{\small
\vspace{-5pt}
\begin{equation}
\label{eq:DefMap5}
\begin{split}
\tilde{m}^{(i,j)} \!=\! m^{(i,j)}-c_{1}(i\!-\!a_{i}\!+\!\frac{c_{3}}{2c_{1}})^2\!-\! c_{2} (j\!-\!a_{j}\!+\!\frac{c_{4}}{2c_{2}})^2,
\end{split}
\vspace{-5pt}
\end{equation}}
\!\!where $m^{(i,j)}$ is the $(i,j)$th element of the part detection map $\mathbf{M}$, $(a_{i}, a_{j})$ is the predefined anchor location of the $p$th part. They are adjusted by $c_3/2c_1$ and $c_4/2c_2$, which are automatically learned.
$c_{1}$ and $c_{2}$ (\ref{eq:DefMap5}) decide the deformation cost. There is no deformation cost if $c_{1}=c_{2}=0$. Parts are not allowed to move if $c_{1}=c_{2}=\infty$. $(a_{i}, a_{j})$ and $(\frac{c_{3}}{2c_{1}}, \frac{c_{4}}{2c_{2}})$ jointly decide the center of the part. 
The quadratic constraint in Eq. (\ref{eq:DefMap5}) can be represented using Eq. (\ref{eq:GenDefMap2}) as follows:
{\small
\vspace{-5pt}
\begin{align}
\tilde{m}^{(i,j)} \! &=\! m^{(i,j)}-c_{1}d_{1}^{(i,j)}- c_{2}d_{2}^{(i,j)}- c_{3}d_{3}^{(i,j)}  \!- c_{4} d_{4}^{(i,j)} \! -\! c_5, \nonumber \\
d_{1}^{(i,j)}\!  &=\! (i-a_{i})^2, \ \ d_{2}^{(i,j)} \!= \!(j-a_{j})^2, d_{3}^{(i,j)}\!  =\! i-a_{i}, \nonumber\\
d_{4}^{(i,j)} &\!= \!j-a_{j}, c_5={c_3}^2/(4c_1)+{c_4}^2/(4c_2).
\label{eq:DefMap6}
\end{align}
 In this case, $c_1, c_2, c_3$ and $c_4$ are parameters to be learned and $d_{n}^{(i,j)}$ for $n=1, 2, 3, 4$ are predefined. $c_5$ is the same in all locations and need not be learned. 
The final output is:
{\small
\vspace{-5pt}
\begin{equation}
\label{eq:DefMap3}
b = \max_{(i,j)} {\tilde{m}^{(i,j)}},
\vspace{-5pt}
\end{equation}}
where $\tilde{m}^{(i,j)}$ is the $(i, j)$th element of the matrix $\tilde{\mathbf{M}}$ in (\ref{eq:DefMap5}).

\begin{figure} 
\centering
\centerline{\epsfig{figure=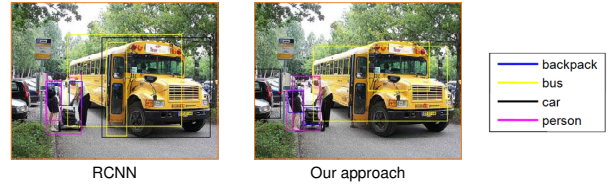,width=8cm}}
\caption{Object detection result for RCNN and our approach. }
\label{Fig:Context}
\end{figure}

\section{Conclusion}
This paper proposes a deep learning diagram that learns four components -- feature extraction, deformation handling, context modeling and classification -- for generic object detection. Through interaction among these interdependent components, the unified deep model improves detection performance on the largest object detection dataset. 
Detailed experimental comparisons clearly show the effectiveness of each component in our approach.
We enrich the deep model by introducing the def-pooling layer, which has great flexibility to incorporate various deformation handling approaches and deep architectures. The multi-stage training scheme simulate the cascaded classifiers by mining hard samples to train the network stage-by-stage and avoids overfitting. The pretraining and model averaging strategies are effective for the detection task. 
Since our approaches are based on baseline deep model, they are complementary to new deep models, e.g. GoogLeNet, VGG, Network In Network \cite{lin2013network}. These recently developed can be used as our baseline deep model to replace AlexNet or Clarifai-fast to further improve the performance of object detection.


\textbf{Acknowledgment}: This work is supported by the General Research Fund sponsored by the Research Grants Council of Hong Kong (Project No. CUHK 417110, CUHK 417011, CUHK 429412) and National Natural Science Foundation of China (Project No. 61005057).

{\small
\bibliographystyle{ieee}
\bibliography{./PME}
}

\end{document}